\documentclass[conference]{IEEEtran}
\IEEEoverridecommandlockouts
\usepackage{cite}
\usepackage{amsmath,amssymb,amsfonts}
\usepackage{algorithmic}
\usepackage{graphicx}
\usepackage{textcomp}
\usepackage{xcolor}
\usepackage{float}
\usepackage{url}

\def\BibTeX{{\rm B\kern-.05em{\sc i\kern-.025em b}\kern-.08em
    T\kern-.1667em\lower.7ex\hbox{E}\kern-.125emX}}
\usepackage{fancyhdr}
\vspace{-.5em}
\vspace{.5em}
\makeatletter
\newcommand*\titleheader[1]{\gdef\@titleheader{#1}}
\AtBeginDocument{%
  \let\st@red@title\@title
  \def\@title{%
    \bgroup\normalfont\small\centering\@titleheader\par\egroup
    \vskip.5em\st@red@title}
}
\makeatother

\title{Lightweight and Scalable Particle Tracking and Motion Clustering of 3D Cell Trajectories}
\titleheader{Accepted to 2019 IEEE International Conference on 
Data Science and Advanced Analytics -- Copyright \copyright 2019 IEEE}
\author{
    \IEEEauthorblockN{Mojtaba S. Fazli\IEEEauthorrefmark{1}, 
    Rachel V. Stadler\IEEEauthorrefmark{3}, BahaaEddin Alaila\IEEEauthorrefmark{1},Stephen A. Vella\IEEEauthorrefmark{2},\\
    Silvia N. J. Moreno\IEEEauthorrefmark{2}, Gary E. Ward\IEEEauthorrefmark{3}, and Shannon Quinn\IEEEauthorrefmark{1}\IEEEauthorrefmark{2}}\\
    \IEEEauthorblockA{\IEEEauthorrefmark{1}Department of Computer Science, The University of Georgia, Athens, GA, USA\\
    \{mojtaba, bahaaeddin.alaila, spq\}@uga.edu}
    \IEEEauthorblockA{\IEEEauthorrefmark{2}Department of Cellular Biology, The University of Georgia, Athens, GA, USA\\
    \{sav28290, smoreno, spq\}@uga.edu}
    \IEEEauthorblockA{\IEEEauthorrefmark{3}Department of Micobiology and Molecular Genetics, University of Vermont, Burlington, VT, USA\\
    \{rachel.stadler, gary.ward\}@uvm.edu}
}

\begin{document}
\maketitle
%


\begin{abstract}
Tracking cell particles in 3D microscopy videos is a challenging task but is of great significance for modeling the motion of cells. Proper characterization of the cell\textquotesingle s shape, evolution, and their movement over time is crucial to understanding and modeling the mechanobiology of cell migration in many diseases.  One in particular, toxoplasmosis is the disease caused by the parasite \textit{Toxoplasma gondii}. Roughly, one-third of the world\textquotesingle s population tests positive for \textit{T. gondii}. Its virulence is linked to its lytic cycle, predicated on its motility and ability to enter and exit nucleated cells; therefore, studies elucidating its motility patterns are critical to the eventual development of therapeutic strategies. Here, we present a computational framework for fast and scalable detection, tracking, and identification of \textit{T. gondii} motion phenotypes in 3D videos, in a completely unsupervised fashion. Our pipeline consists of several different modules including preprocessing, sparsification, cell detection, cell tracking, trajectories extraction, parametrization of the trajectories; and finally, a clustering step. Additionally, we identified the computational bottlenecks, and developed a lightweight and highly scalable pipeline through a combination of task distribution and parallelism. Our results prove both the accuracy and performance of our method.
\end{abstract}

\begin{IEEEkeywords}
Large scale 3D Cell Tracking, Motion Trajectories, Geodesic distance, Spectral Clustering, \textit{Toxoplasma gondii}, Martin Distance
\end{IEEEkeywords}

\section{Introduction}
\textit{Toxoplasma gondii} is one of the most successful intracellular parasites and is the causative agent for toxoplasmosis: a disease that afflicts roughly one-third of the world\textquotesingle s population \cite{b1}. For infants, elderly, neonatal fetuses, and individuals with weak or compromised immune systems, \textit{T. gondii} can cause deadly complications \cite{b1, b2}. It invades the cells, hijacks the cellular machinery and replicates, before egressing and beginning subsequent rounds of invasion on new host cells. The lytic cycle is the whole process of motility, invasion, replication, and egress.  The disease progression and virulence of \textit{T. gondii} is a direct consequence of the tissue destruction of thousands rounds of successful lytic cycles events. \textit{T. gondii}\textquotesingle s lytic cycle is entirely predicated on the parasite\textquotesingle s motility, thus underlying the significance of studying the motion of \textit{T. gondii}.

We seek to model the motion of \textit{T. gondii}. Doing so will help address some of the underlying questions regarding \textit{T. gondii} mechanobiology: what the fundamental motion phenotypes of \textit{T. gondii} are, and how they are coordinated to enable for a successful lytic cycle, and how they may indicate the internal parasite state. Using pharmacology, we can study the molecular basis of the cell signaling pathways that regulate parasite motility, and track how it relates to the parasite\textquotesingle s motility. Ideally, our goal will assist us in the development of future interference strategies against \textit{T. gondii}\textquotesingle s lytic cycle.

The goal of our project is to model the motion of \textit{T. gondii}. In order to achieve this goal, we need to track the cells and quantify the cell motion and its spatial features across time. There are several cell tracking models that are proposed for tracking both in 2D and 3D videos. Some of them are suggested for cell tracking \textit{in vivo} and some others for tracking the cells in a laboratory environment (\textit{in vitro}). Hirose et al. \cite{b6} proposed a motion model with a Markov Random Field to track hundreds of cells in 4D live-cell imaging data. In another study \cite{b7}, the researchers proposed a model that uses GMMs for segmentation and Kalman filters for tracking the spermatozoa. In another study, Mingli Lu et al. \cite{b8} tracked multiple cell objects in 2D space. They used multiple particle filtering methods to track multiple cells with a variety of collisions and divisions among the cells. Moreover, researchers have used some conventional methods for tracking without principal reliance on computer vision: “Physical Tracking Method,” which means moving the sample automatically to keep one individual cell in focus \cite{b11, b12}, is one of these kinds of methods. However, there are some limitations for using this method, including the requirement of a specific experimental setup, and its restriction on observing multiple cells at a time. In another method \cite{b13}, the researchers moved the focus along the depth axis, and then used image cross-correlations, compared the observed patterns with a library of reference images, and assigned a $z$ position based on the identity of the best match. Both methods use only first-order image features to perform tracking and are thus susceptible to even mild sources of noise or perturbation.

Previous work mostly concentrated on 2D videos and developed some pipelines for detection, tracking and parameterization of 2D cell motion\cite{b3}. An unsupervised method for grouping the latent motion phenotypes of the cell in 2D space was proposed in \cite{b4}. Motion parameterization is not a new concept, others have used it in different research areas previously\cite{b26, b17, b4}. Although the results in 2D space study were satisfactory, there were some shortcomings. In \cite{b4}, the authors used an RBF (Radial Basis Function) kernel for computing the distance in a clustering pipeline. However, since the parameterized trajectories are living in a geodesic space, the RBF kernel could not account for the nonlinearities in the latent space. Second, these studies were conducted on 2D microscopy data; new research has suggested that the motion of \textit{T. gondii} in a 3D space differs qualitatively from the observed motions in a 2D space \cite{b14}. Whether this is an artifact of different experimental conditions or a real biological phenomenon remains unclear. Nevertheless, using experimental conditions that more closely resemble \textit{in vivo} circumstances--i.e., parasite motilities in 3D space--is always preferable and may open new horizons to the researchers in this field.

On one hand, tracking 3D objects may help researchers overcome some limitations of 2D microscopy, such as occlusions and out-of-focus objects. On the other hand, 3D datasets are challenging to acquire and analyze. Crucially, 3D data is significantly larger in size, increasing the computational costs; moreover, it raises the degrees of freedom in object tracking, and as a consequence magnifies the computational complexity. Ultimately, it seems that studying 3D video microscopy could be more beneficial, as it reflects reality much better than 2D videos. To fully leverage this modality, one must look for new computational approaches that can scale to the data adequately. Here, we propose to build on our conceptual advances from our previous works in parallel and distributed analysis of 4D functional MRI brain scans\cite{b9, b10} using Apache Spark. Here, we use Dask\cite{b28}, a hihgly parallel and distributed task-manager, in order to achieve a high level of concurrency.

In summary, we developed a computational pipeline for tracking \textit{T. gondii} parasites in 3D microscopy videos. Our pipeline consists of 6 different modules, including: preprocessing, cell detection, cell tracking, trajectory extraction, parametrization of the trajectories, and finally a motion-parameters clustering module. After developing the serial version of our computational framework, we provided a scalable version using the Dask framework. Finally, we tested the pipeline on the Google Cloud Platform (GCP) with a different number of workers and conclude with performance comparisons between the serial and the distributed versions of our pipeline.

\section{Data}
In this study, our dataset consists of twelve videos with a total size of $24 GB$. Each video includes around $63$ frames; each frame has $41$ spatial slices (along the $z$-axis), and each slice is a $500\times502 $ grayscale image. The slices are recorded in raw format as RGB TIFF images indexed by $z-axis$ (depth) and $t-axis$ (time). These videos are obtained from different experiments with different number of cells.

\begin{figure}
\centerline{\includegraphics[width=9cm]{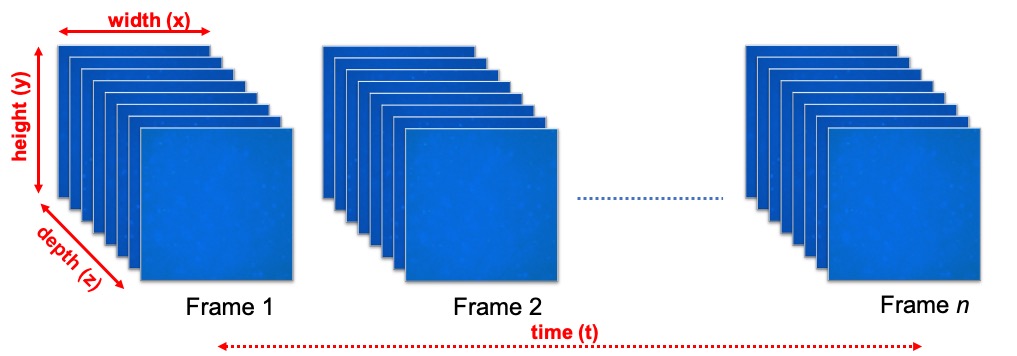}}
\caption{\textit{3D microscopy video slices arrangement in different frames across the video.}}
\label{fig:1}
\end{figure}

\begin{figure*}
\centering
\includegraphics[width=1.0\textwidth]{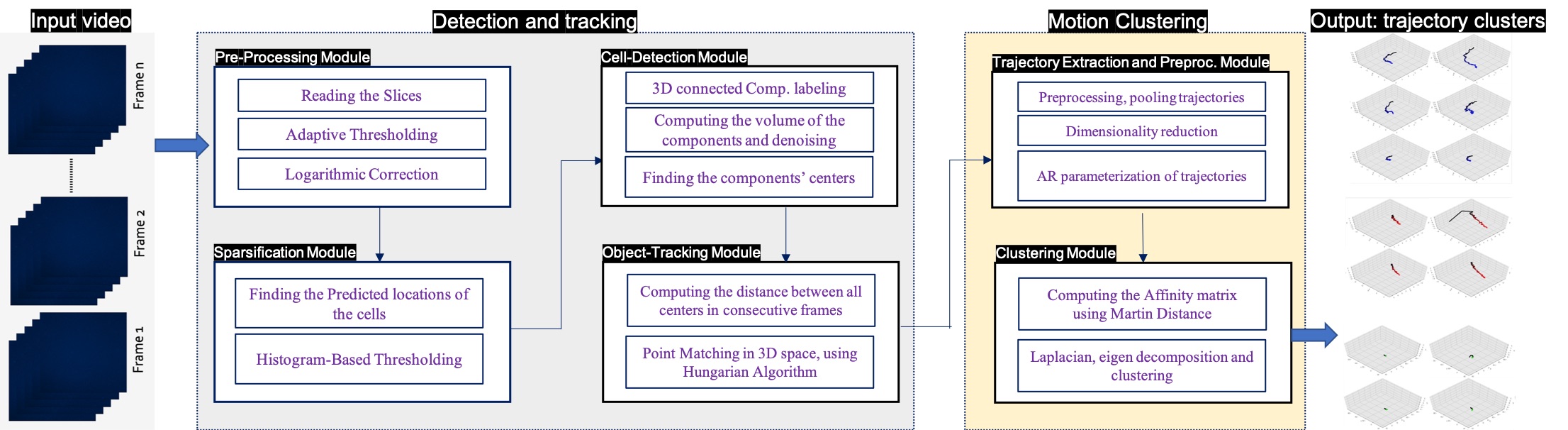}
\caption{\label{fig:2} \textit{General overview of our computational platform: the input video passes through preprocessing and sparsification first to highlight the cell particles in the image slices. Then through cell detection and tracking the cell particles are captured, grouped as cells, and tracked across time-frames. The resulting trajectories get preprocessed, pooled, get their dimensionality reduced, AR-paramterized, then clustered.  }}
\end{figure*}

This data was acquired using a PlanApo $20X$ objective ($NA = 0.75$) on a preheated Nikon Eclipse $TE300$ epifluorescence microscope. Time-lapse stacks were captured using an $iXon885$ $EMCCD$ camera (Andor Technology, Belfast, Ireland) cooled to $-70^oC$ and driven by NIS Elements software (Nikon Instruments, Melville, NY).  The camera was set to frame transfer sensor mode, with a vertical pixel shift speed of $1.0 µs$, vertical clock voltage amplitude of $+1$, readout speed of $35 MHz$, conversion gain of $3.8\times$, EM gain setting of $3$ and $2×2$ binning. The $z-slices$ were imaged with an exposure time of $16$ ms; stacks consisted of 41 $z-slices$ spaced $1 µm$ apart for a total $x, y, z, t$ imaging volume of $402 \mu m \times 401\mu m \times{40} \mu m$ for $62$ stacks in $60 s$. All experiments were completed within $80$ min of harvesting the parasites. Fresh parasites were harvested both at the very beginning and end of each imaging experiment to ensure imaging and parasite conditions remained constant \cite{b14}.

\section{Computational Framework}
\subsection{Software}
We implemented our pipeline using Python $3.6$ and associated scientific computing libraries (NumPy, SciPy\cite{b30}, scikit-learn\cite{b15}, matplotlib). The core of our detection and tracking algorithm used a combination of tools available in the OpenCV $3.1$ computer vision library \cite{b16}. For the distributed version of the core platform, we used Dask Arrays and Dask dataframes.
Dask-ml and Dask-ndarrays were used to implement some of the machine learning tasks in the distributed version of our pipeline.  For local parallelization and multiprocessing, we used the joblib backend and the multiprocessing library of Python. The full code for the proposed framework is open sourced on github under the MIT open source license at \textit{\url{https://github.com/quinngroup/3Dcell_tracking_DSAA2019}}.

\subsection{Problem Definition}
A crucial step towards understanding and modeling the motion model of \textit{T. gondii} is capturing the cell motions at relevant spatial and temporal scales. In 3D microscopy videos, each frame has an additional depth dimension. The microscope\textquotesingle s camera achieves a fast acquisition of the slices across different depths, though at the expense of quality: the images are very noisy. 

The acquisition of the 3D microscopy slices across time is demonstrated in Fig. \ref{fig:1}. Slices are the 2D images with size of $500\times 502$ pixels, forming cross sections of depth in the 3D video. Our first step was to detect the partial spatial locations of the cells in each slice of a single frame, then unify those discrete particles and compute the center of the detected mass across all $z$-slices of one frame. Finally, we tracked those 3D particles across time.
After tracking all objects in all the videos, we created a giant trajectory pool, parametrized them, and clustered the movement patterns. 
\newline
\begin{figure*}[h!]
\centering
\includegraphics[width=1.0\textwidth]{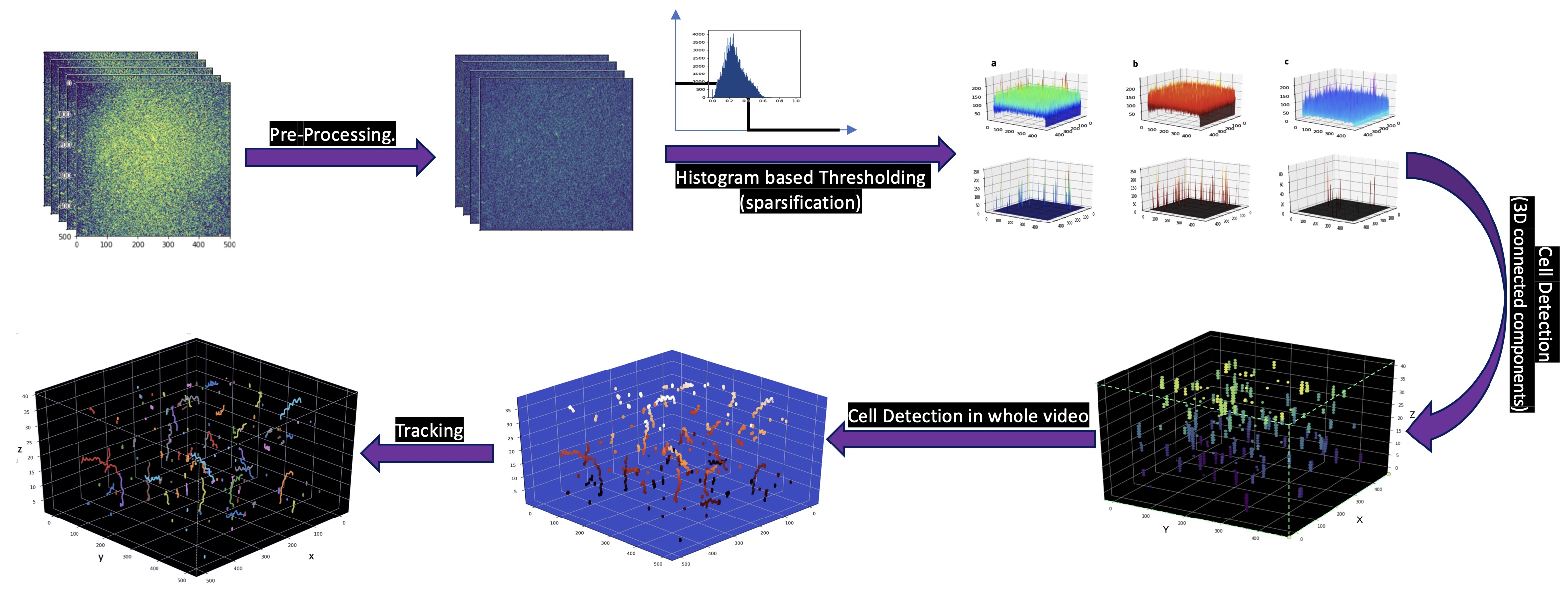}
\caption{\label{fig:3} \textit{The detection \& tracking parts in our framework: the input to this part is a set of 2D images as different slices along the depth of each time-frame, and the output is a data structure in which we have all the trajectory points across the time-frames. All these steps are explained in our previous work \cite{b5} in full details}}
\end{figure*} 
\begin{figure}
\centerline{\includegraphics[width=9cm, height=3.5cm]{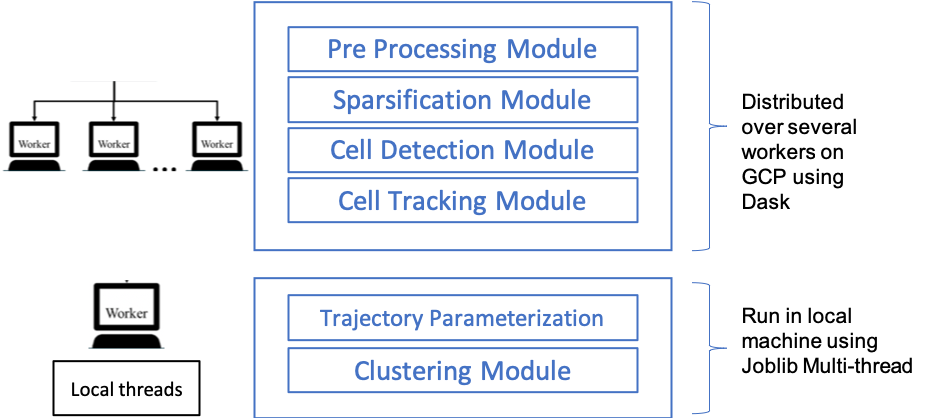}}
\caption{\label{fig:4} \textit{schema of the proposed distributed framework}}
\end{figure}
\subsection{Model}
We developed the computational pipeline both in serial and later as a distributed engine. Fig. \ref{fig:2} illustrates the serial version of our framework, showing a general view of the architecture, its modules, and the components of each module. The pipeline consists of two different parts: the detection \& tracking part and the motion clustering part. The detection \& tracking component includes four different modules. First, the pipeline reads the input data. The input data arrangement is in the form of image slices.  In general, our data consists of multiple 4D matrices with the following dimensions:

\begin{equation}\label{1}
	f(t, z, x, y) \in \mathbb{R}^{63 \times 41 \times 500 \times 502}
\end{equation}

In \eqref{1} $f$ indicates a video, $x$, $y$ and $z$ are spatial dimensions and $t$ is the temporal dimension. In our datasets, the last three dimensions of eq. \eqref{1} are constant (the $x$,$y$, and $z$ dimensions are the same for all the videos), but the number of frames varies among different videos.

\subsection{Cell Detection and Tracking}

Fig. \ref{fig:3} illustrates the cell detection and tracking part of our pipeline. After reading the 3D videos, we converted them into numpy arrays of grayscale images. Next, we preprocessed our data. After that, our pipeline created a sparse matrix containing the cell particles across the $z$-slices. Then, we consolidated the particles across depth to materialize the cells in 3D space. Finally, the pipeline tracked the objects across time by minimizing the cost between the points in consecutive frames using the Hungarian algorithm\cite{b29}. The serial version of our tracking pipeline and its components are discussed in detail in\cite{b5}. Here, we focus on the second part of our framework: the improvements that we made to the pipeline in order to have a fast and distributed framework. The distributed schema of the framework is shown in Fig. \ref{fig:4}. As it indicates, our parallel version is divided into two sections. We explain this framework in the following sections.

\subsection{Preprocessing and normalization of the trajectories}

The cell detection and tracking part of the framework extracts the trajectories for each video. These trajectories are comprised of a 3D position vector for each detected particle in each frame. As a side effect of thresholding and denoising, sometimes the length of the trajectories were not equal. Moreover, since the number of frames in different videos varied between $61$ and $63$ frames, we needed to make these trajectories of equal length to be used in downstream analysis. Thus, we discarded incomplete trajectories (below 61 frames) and truncated longer trajectories to be 61 frames.

After applying this preprocessing step, we ended with a corpus of approximately 3,000 complete trajectories, each one indicating the 3D movement of a single cell.  Our data was represented in the following structures, where $X$, $Y$, and $Z$ represent $x$, $y$ and $z$ coordinates of the objects over time; these matrices are demonstrated in Eqs. \eqref{2}-\eqref{4}.

\begin{equation}\label{2}
    X_{(3000\times61)} = \begin{bmatrix} 
    x_{(0,0)} & \dots & x_{(0,60)} \\
    \vdots & \ddots &  \vdots \\
    x_{(2999,0)} & \dots & x_{(2999,60)} 
    \end{bmatrix}
\end{equation}
\begin{equation}\label{3}
    Y_{(3000\times61)} = \begin{bmatrix} 
    y_{(0,0)} & \dots & y_{(0,60)} \\
    \vdots & \ddots &  \vdots \\
    y_{(2999,0)} & \dots & y_{(2999,60)} 
    \end{bmatrix}
\end{equation}
\begin{equation}\label{4}
    Z_{(3000\times61)} = \begin{bmatrix} 
    z_{(0,0)} & \dots & z_{(0,60)} \\
    \vdots & \ddots &  \vdots \\
    z_{(2999,0)} & \dots & z_{(2999,60)} 
    \end{bmatrix}
\end{equation}
\begin{figure*}[h!]
\centering
\includegraphics[width=1.0\textwidth]{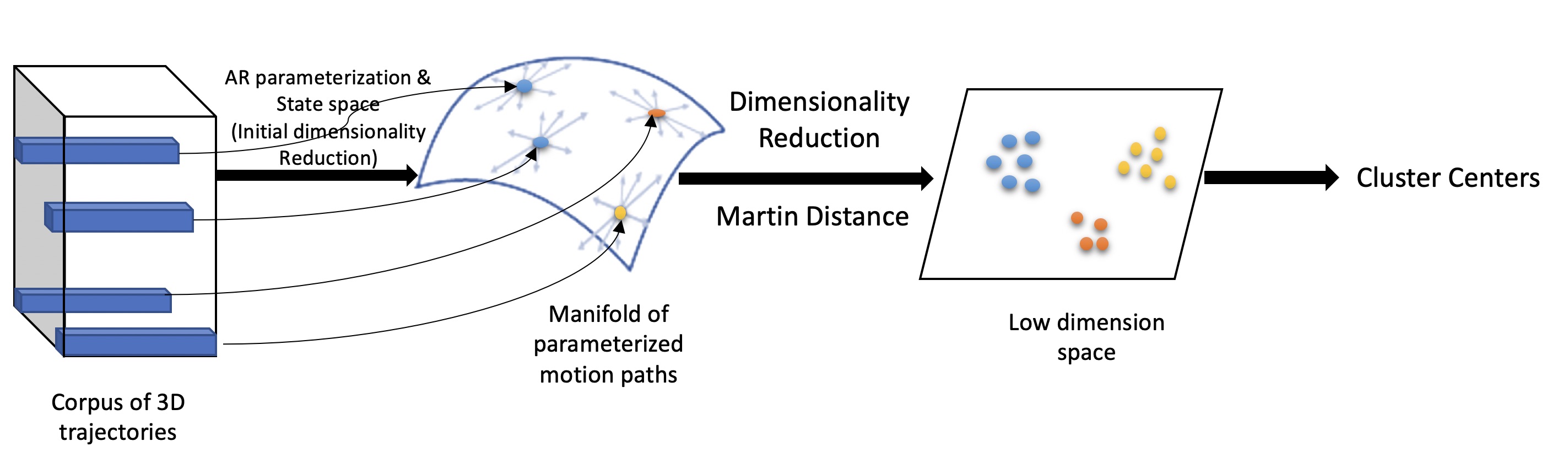}
\caption{\label{fig:5} \textit{The parameterization and clustering model in our pipeline. We create a corpus of trajectories, then we parameterize our trajectories with AR transition matrices and after applying dimensionality reduction. As a result a manifold of parameterized motion paths is created. Then we use Martin distance and spectral clustering to cluster the trajectories }}
\end{figure*}
\subsection{Parameterization of trajectories}

In order to perform unsupervised discovery of \textit{T. gondii} motility phenotypes, we featurized the trajectories of each cell such that the features were invariant to absolute spatial coordinates, but remained sensitive to relative changes in position over time. Therefore, we used a linear dynamical system that encodes a Markov-based transition prediction method: an autoregressive (AR) model. We used the formulation of AR processes as defined in \cite{b18, b19, b20}. The complete AR model is specified as follows:
\begin{equation}
\vec{y}_t = C\vec{x}_t + \vec{u}_t\label{5}
\end{equation}

\begin{equation}
\vec{x}_t = B_1\vec{x}_{t-1} + B_2\vec{x}_{t-2} + ... + B_d\vec{x}_{t-d} + \vec{v}_t\label{6}
\end{equation}
where $\vec{r}_t \in \mathbb{R}^{p}$ in Eq. \eqref{5} denotes a particle\textquotesingle s position in frame $t$ and $p$-coordinates, here $p$=3 --the three spatial coordinates, $\vec{r}_t=(x_t,y_t,z_t)$. Eq. \eqref{5} factorizes the position of a particle into a noise vector $\vec{u}$, and a lower-dimensional hidden state ${\vec{h}_t} \in \mathbb{R}^n$ ($n<p$) projected back into $p$-coordinates using a projection matrix $C \in \mathbb{R}^{p \times n}$. All particle trajectories share $C$, ensuring that all hidden states $\vec{h}_t$ live in the same latent space.  Eq. \eqref{6} denotes an Markov system of $d$-order and represents the dynamics of the hidden state $\vec{h}$ of the \textit{T. gondii} motion in the latent subspace defined by the projection matrix $C$. In other words, ${\vec{h}_t} \in \mathbb{R}^n$ is the hidden state for position at time $t$ and the matrices $B_i\in \mathbb{R}^{n\times n}$ represent the dynamics of the system given the hidden states in the previous $d$ frames. $\vec{u}$ and $\vec{v}$ denote the residuals in the original and latent spaces respectively. To perform the motility parameterization, we are interested in the transition matrices $B_i$, which are quantitative encodings of \textit{how the system evolves}, and therefore represent an encoding of the motion that is robust to absolute spatial positions.

Since our trajectories live in a $3D$ space in each frame, the dimensions of each trajectory are $3\times{t}$. And for $m$ trajectories, the total size of the corpus ($T$) will be a matrix with the following size:
\begin{equation}\label{7}
	T \in \mathbb{R}^{3 \times m \times t}
\end{equation}

First, we use PCA(Principal Component Analysis) to obtain a projection matrix $C \in \mathbb{R}^{3 \times 2}$. In other words, we map our $3D$ space into a $2D$ embedding space. As a consequence, the new dimensions of our latent trajectories matrix $H$ will be of size ${2\times{m}\times{t}}$. In our study, we found that a system with the order of $d = 5$ works well. Upon computing the AR parameters $\big\{B_1...B_5\big\}$, each trajectory is represented as five $2\times{2}$ matrices(or flattened, a 20-dimensional vector). And for $m$= 3,000 trajectories, the $AR$ parameter corpus size is of the following shape:

\begin{equation}\label{9}
	AR_{Mat} \in \mathbb{R}^{(2 \times 2 ) \times 5 \times 3000}
\end{equation}

\begin{equation}{\label{8}}
    AR_{(3000\times20)} = \begin{bmatrix} 
    AR_{(0,0)} & \dots & AR_{(0,19)} \\
    \vdots & \ddots &  \vdots \\
    AR_{(2999,0)} & \dots & AR_{(2999,19)} 
    \end{bmatrix}
\end{equation}
\subsection{Establishing pairwise motion similarities }
The $AR$ motion parameters are strong quantitative representations of movement that are invariant to absolute spatial coordinates \cite{b4}. However, they live in a geodesic space that does not span a Euclidean space \cite{b21, b22}, and consequently, are not amenable to analysis methods such as K-means that rely on Euclidean-based pairwise distance metrics. Thus, we constructed a custom kernel matrix using Martin distance \cite{b22}.
\textit{Martin distance} is a geodesic non-Euclidean unbounded distance metric that is constructed based on the subspace between two linear dynamical systems. Martin distance is effective for measuring the similarity of dynamical systems like AR models \cite{b20}. Martin distance is defined over both the motion parameters $B$ and the subspace $C$:

\begin{equation}
	B^{T}PB = -C^{T}C\label{10}
\end{equation}
For calculation of subspace angle\cite{b20}, we need to solve the above Lyapunov equation \eqref{10} for $P$. Where for two trajectories of $t_i$ and $t_j$ and some number of subspace dimensions $q$ we have:  

\begin{equation}\label{11}
    P = \begin{bmatrix} 
    P_{11} & P_{12} \\
    \\
    P_{21} & P_{22} 
    \end{bmatrix} \in \mathbb{R}^{2q\times 2q}
\end{equation}

\begin{equation}\label{12}
    B = \begin{bmatrix} 
    B_{t_i} & 0 \\
    \\
    0 & B_{t_j} 
    \end{bmatrix} \in \mathbb{R}^{2q\times 2q}
\end{equation}
\\
\begin{equation}\label{13}
    C = \begin{bmatrix} 
    C_{1}  & & C_{2} 
    \end{bmatrix} \in \mathbb{R}^{p\times 2q}
\end{equation}
By performing eigendecomposition over the symmetric matrix of $P$ we compute the angle between the subspaces as:
\begin{equation}\label{14}
	\cos^2{\theta_i} = i^{th}eigenvalue(P_{11}^{-1} P_{12}^{} P_{22}^{-1} P_{21}^{})
\end{equation}
Finally, by using \eqref{14} and finding the eigenvalues we can compute the Martin distance between two trajectories as:

\begin{equation}\label{15}
d_{M}(t_i, t_j)^2 = -ln\prod_{k=1}^{q} \cos^2\theta_k
\end{equation}
In our application, $C_1$ and $C_2$ are equal since we parameterized all trajectories using the same matrix $C$ in Eq.\eqref{5}. Using equations \eqref{10}-\eqref{15}, we compute the pairwise Martin distance of the parameterized trajectories. The goal here is to compute a similarity matrix between the trajectories to use for spectral clustering. However, the Martin distance is a divergence and not an affinity(similarity). Hence, to have the Martin affinities, we use the RBF (heat kernel) \eqref{16} on Martin distance values, (i.e. substituting the Euclidean distance in regular RBF kernels with the Martin distance for our setting):

\begin{equation}\label{16}
S_i = e^{\frac{{-\beta}\times{M_i}}{\sigma}}
\end{equation}
where, $\beta$ is a free parameter and $\sigma$ is the standard deviation of  Martin affinities $M$.

\subsection{Spectral clustering to identify similar motion phenotypes} \label{clusteringsection}

\begin{figure}
\centerline{\includegraphics[width=0.5\textwidth]{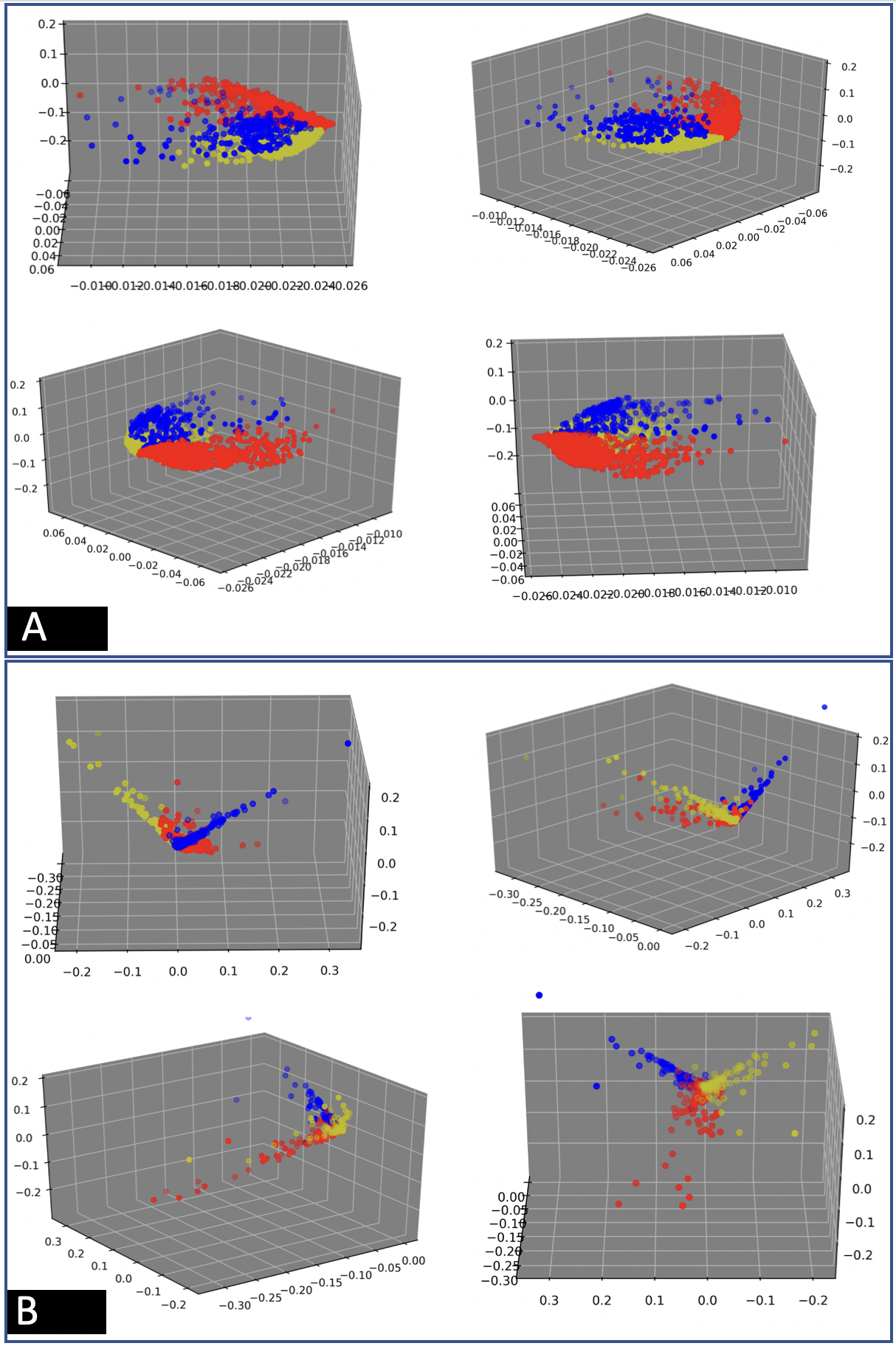}}
\caption{\textit{\textbf{a:} Spectrum of motion visualization for around 3000 collected parameterized trajectories from different videos plotted in different angles (90,180,270,360). Each dot represents a parameterized cell trajectory in the lower-dimensional space. The colors indicate the different clusters as reported by spectral clustering. \textbf{b:} The same visualization for a specific video with around 640 trajectories. Each subplot shows the same data, with the viewing angle rotated 90 degrees.
}}
\label{fig:6}
\end{figure}

Clustering methods are generally used for unsupervised learning; they typically impose few assumptions on the data, defining only the notion of similarity with which to group the data. K-means is a favorite clustering method for data scientists because of its speed and simplicity. However, in our case, the drawbacks of K-means and other conventional methods make these methods an impractical option.

Firstly, K-means assumes Euclidean distances; AR motion parameters are geodesics that do not reside in a Euclidean space. Secondly, K-means assumes isotropic clusters; although AR motion parameters may exhibit isotropy in their space, without a proper distance metric, this issue cannot be clearly examined.
\begin{figure*}[h!]
\centering
\includegraphics[width=1.0\textwidth]{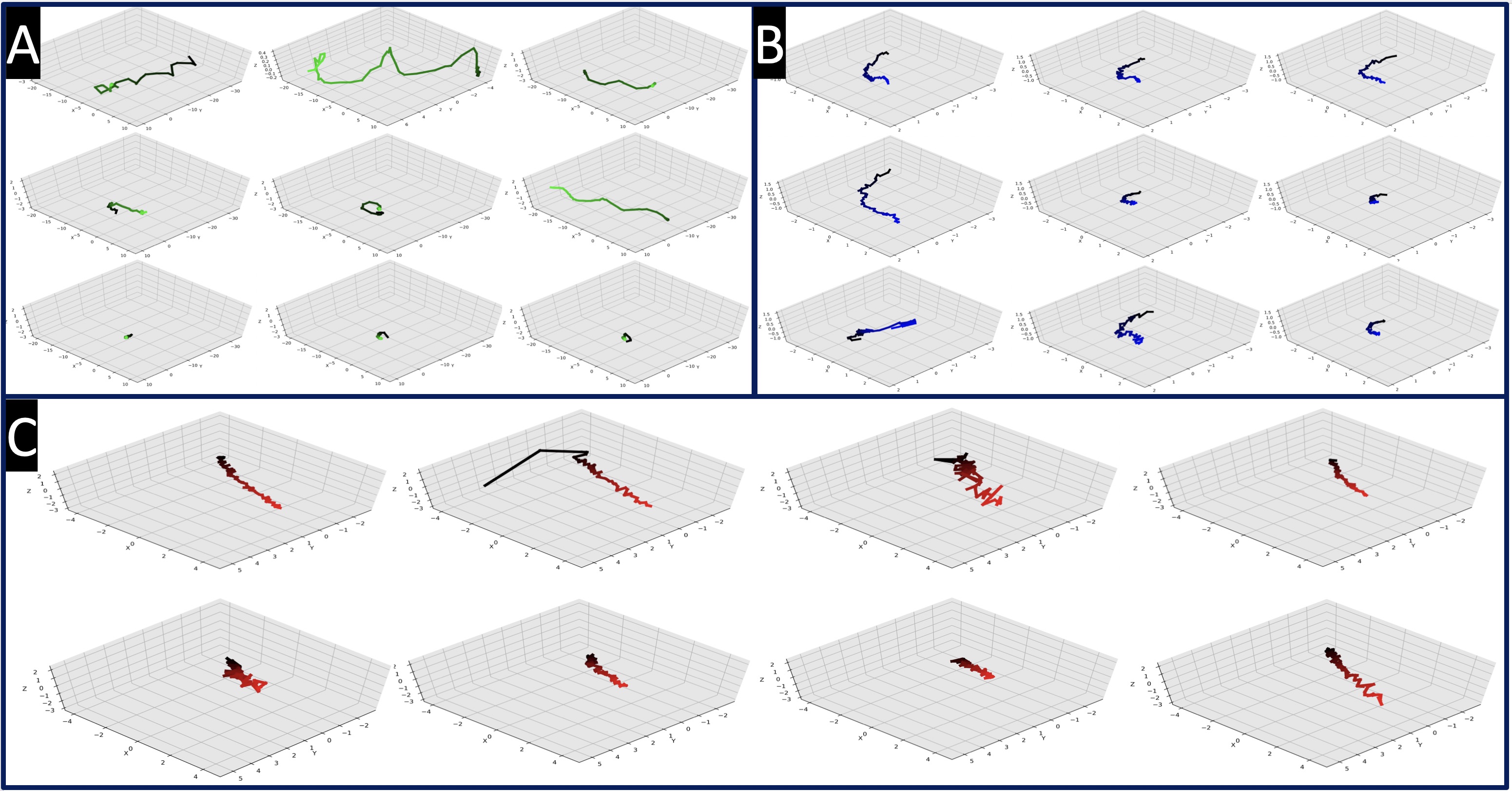}
\caption{\label{fig:7} \textit{Results of trajectory clustering: there are 3 specific clusters A, B and C. Samples from each cluster are visualized in 4D subplots. Different colors indicate different clusters. In each 4D sample plot, we visualize the 3 spatial dimensions, as well as the temporal dimension indicated by the hue in each sample (darkens as time progresses).}}
\end{figure*}
\begin{figure}
\centerline{\includegraphics[width=9cm, height=5.5cm]{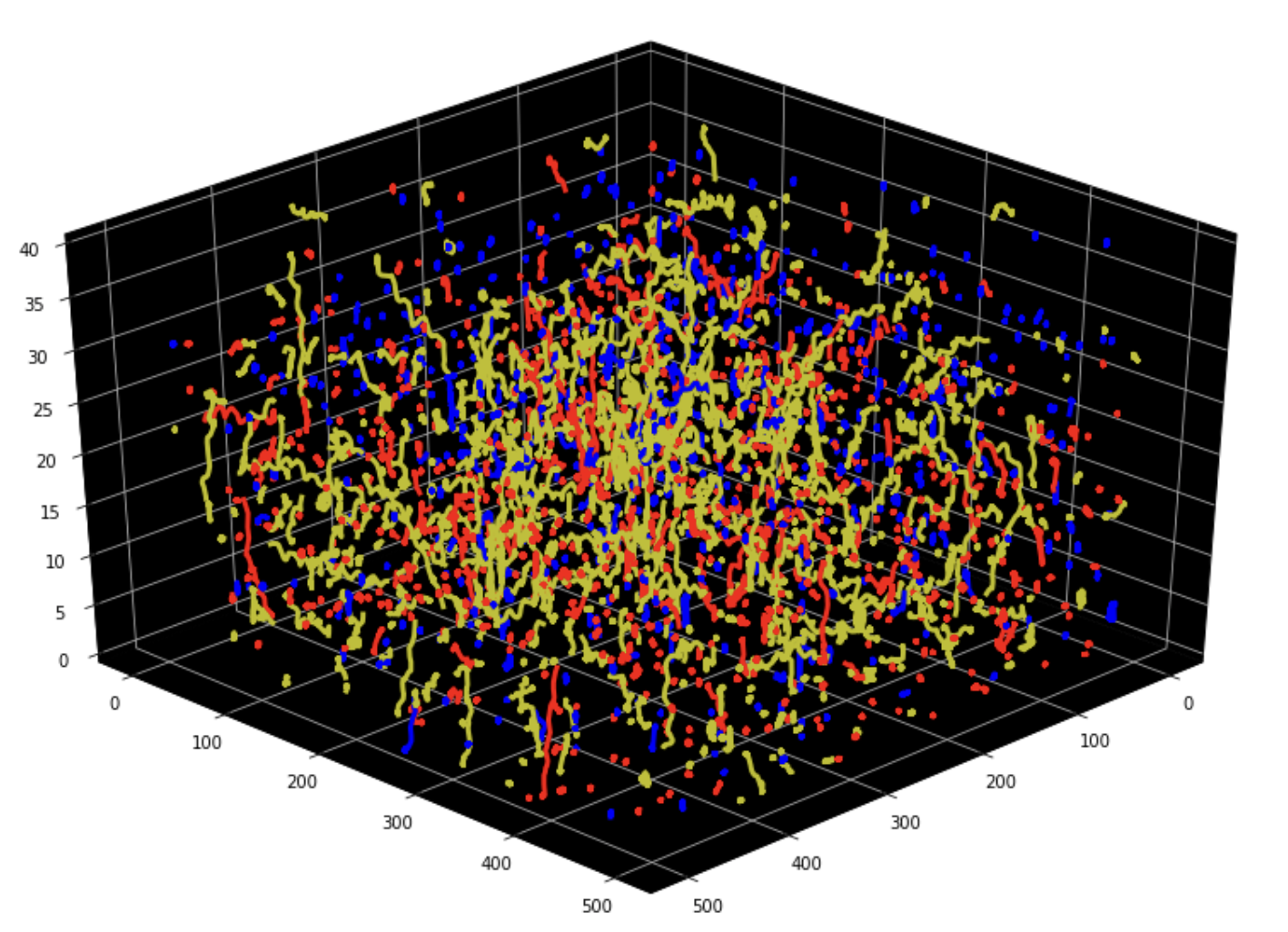}}
\caption{\textit{Aggregated plot of all trajectories colored by their cluster number (3 clusters: red, blue and yellow).}}
\label{fig:8}
\end{figure}
Therefore, we choose spectral clustering. 

Spectral clustering works on the spectrum of the underlying graph of the data \cite{b23}, imposed through a similarity computation using a pairwise kernel. Using the Martin affinity kernel discussed above, a graph Laplacian of the Martin similarity matrix is computed. Then the eigenvectors of the affinity matrix are used to embed the original data in a low-dimensional space where they are separable by a simpler clustering strategy such as K-means (on the eigen-embedding space).  

This requires a full diagonalization of the Laplacian matrix and, therefore, can pose a computational bottleneck with large data. In this study, our 3,000-trajectory corpus was manageable. Computing the Martin affinities between pairwise trajectories (Eqs. \eqref{10}-\eqref{16}) posed a computational bottleneck that scaled quadratically with the number of trajectories. However, using the joblib backend allowed us to parallelize this step locally . 

We used an out-of-the-box spectral clustering implementations in scikit-learn \cite{b24}. Fig \ref{fig:5} depicts the process of parameterizing and clustering the motion trajectories. It is worth mentioning that, the Dask-ml library has a version of fast spectral clustering which uses the Nystr{\"o}m Method as an approximation for extracting the eigenvectors without diagonalizing the large $m \times m$ similarity matrix\cite{b25}. However, the Nystr{\"o}m Method that is used by Dask-ml is not an accurate approximation and requires a large sample for the basis of the eignvectors, therefore we decided against using it.

\section{Results}

By studying and examining the eigenvalues of the affinity matrix computed through the martin distance, and verifying the plotted affinity matrix, we chose $k = 3$ clusters for our analysis (a drop in the eigenvalues after 3).

\subsection{Spectrum of motion phenotypes}

We visualized the spectrum of motion, Fig. \ref{fig:6}. After computing the graph Laplacian $L$ from the affinity matrix of Martin distances, we used the three largest eigenvectors of $L$ to embed each trajectory in a low-dimensional subspace spanned by the principal components of $L$, and applied K-means clustering on this embedding. The colors indicate the cluster label for trajectory as reported by the K-means step.

We found that, in each video, the spectrum of motion is more distinct than the patterns obtained when we pool all the trajectories from all videos. More specifically, by comparing Fig. \ref{fig:6}a and Fig. \ref{fig:6}b, one can see that when all the trajectories are processed together (Fig. \ref{fig:6}a), the structure behind the motion trajectories that exist in individual videos, like (Fig. \ref{fig:6}b), is undermined to some extent.
 There may be some video-specific artifacts or other conditions that exist exclusively in individual videos to explain these results.

\begin{figure*}[h!]
\centering
\includegraphics[width=1.0\textwidth]{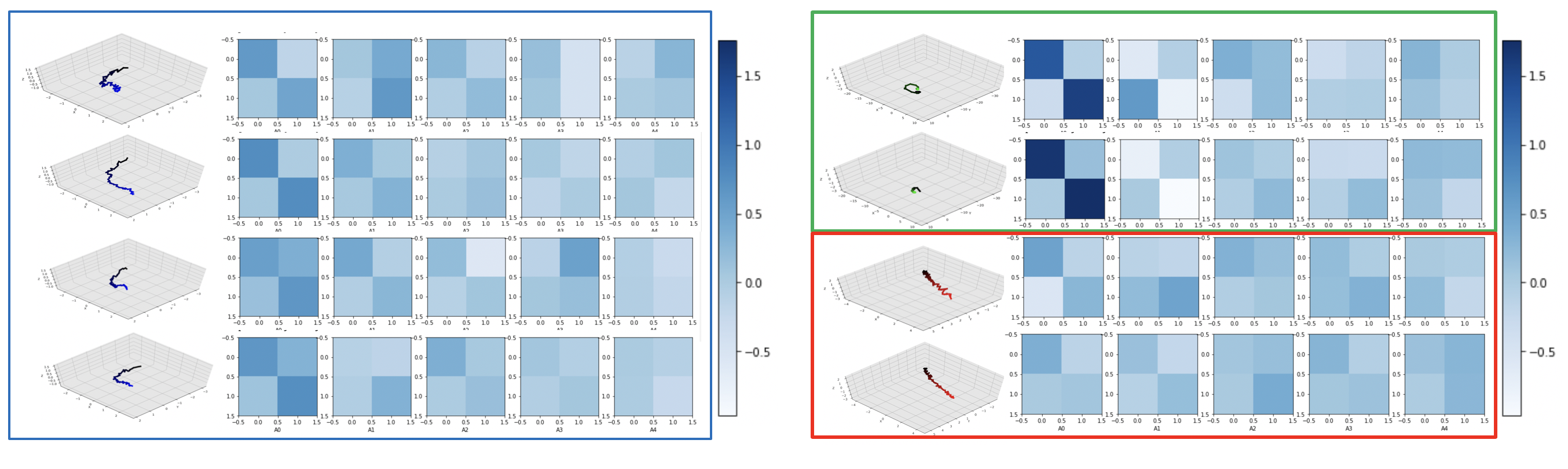}
\caption{\label{fig:9} \textit{AR parameters (transition matrices) visualization for sample trajectories from each cluster. The sample trajectories are visualized together with their corresponding five 2x2 transition matrices. Samples from different clusters are colored differently}}
\end{figure*}
\subsection{Visual inspection of the clusters}

Fig. \ref{fig:7} shows a summary of the clustering results. Each subplot indicates a specific cluster. Each 3D sub-figure is a single object trajectory, colored by a line, showing the extent of the object\textquotesingle s movement over time. The hue indicates the passage of time, with the lightest point of each color indicating the first frame, and the darkest point denoting the last frame.

We can observe the unmistakable similarity of trajectory dynamics within each cluster. Cluster $A$ indicates a full or partial circular motions; there is a helical or circular motion in each of the sample trajectories plotted in this subplot. In cluster $B$, we see an erratic semi-circular motion that progresses in a clockwise direction (in the plotted point of view). Finally, cluster $C$ shows a corkscrew-shaped motion in a straight line. Thus, the motilities in each cluster are semantically different from each other, especially when accounting for the temporal dimension in each trajectory. Moreover, when viewed in context, (Fig.\ref{fig:8}), with more scrutiny, one can identify some distinct \textit{T. gondii} motility phenotypes.
This strongly suggests that there is more than one motion pattern for \textit{T. gondii} in 3D space, contrary to previous suggestions \cite{b14}. In other words, this shows that our results on 3D are in line with prior \textit{T. gondii} research in the 2D setting\cite{b31,b4}, which at the very least means more work on 3D \textit{T. gondii} microscopy analysis is needed, or perhaps there, indeed, are biologically-relevant discrete motion phenotypes that can be indicative of lytic cycle changes. 

We also visualized the AR transition matrices of some sample trajectories from each cluster in Fig. \ref{fig:9}. The significance of the transition matrices resides in the fact that they encode the system evolution over time quantitatively. As the figure demonstrates, while the transition matrices of the trajectories within a cluster are quite similar, there are some visible differences among the AR matrices of the trajectories of different clusters. These differences underline the meaningful clustering of the trajectories.

\section{Distribution and Parallelization}

Although our computational pipeline was designed with efficiency in mind, it still suffers from bottlenecks, primarily because of the size of the dataset. Therefore, we began building in parallelization and/or distribution of the computation. In the serial version of our pipeline, we repeat most of the steps for multiple slices and multiple frames in the different modules. Many of these computations are independent and can be done separately, before feeding the extracted particles into the tracking module. For the sake of scalability and faster computational speed, we used Dask, in addition to its out-of-the box compatibility with the rest of the tools in our pipleline (Numpy, Pandas, scikit-learn).  

Dask\cite{b28} is a distributed scheduler for large scale data processing. It is a Python framework that utilizes a joblib backend locally, and a master-slave model across cluster nodes. To use Dask, we needed to construct our pipeline using Dask\textquotesingle s data structres and operations instead of numpy ones. The resulting pipeline can be run either on a local machine or across a cluster like Google Cloud Platform (GCP) or Amazon Web Services (AWS).

\subsection{Dask version of our pipeline}
We uploaded the dataset to Google Storage. This allowed all dask workers to have access to the data in parallel using dask distributed read functionality through Google Cloud File System (\textit{gcfs}) API.  We read the data into Dask-arrays chunked frame-wise (i.e. guaranteeing that slices of the same time-frame end up on the same worker).

\begin{figure}
\centerline{\includegraphics[width=9cm, height=4cm]{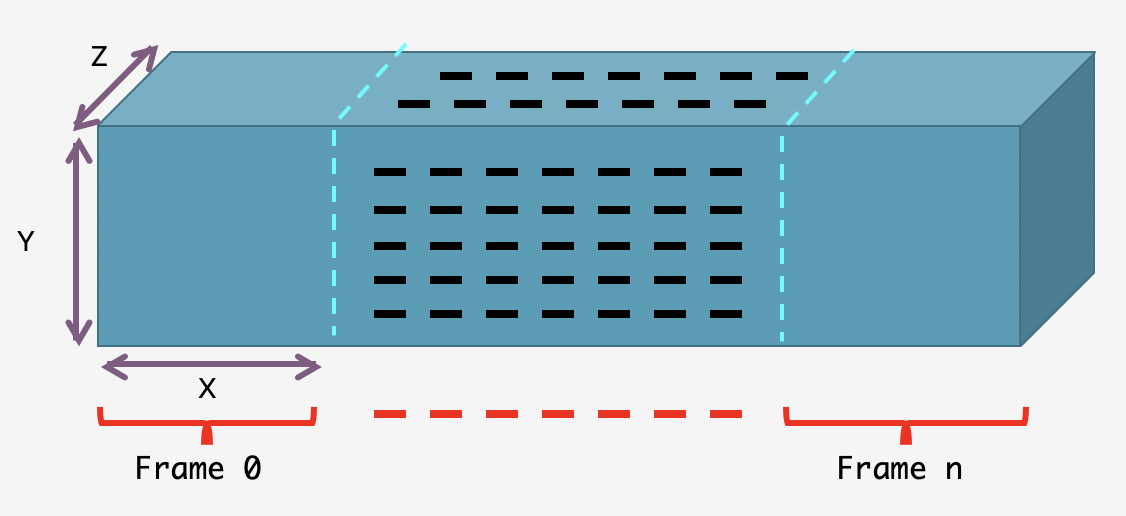}}
\caption{\textit{The structure of the Dask-array chunked frame-wise. Each time-frame ends up on the same worker.}}
\label{fig:10}
\end{figure}
For the clustering part of our framework, since we had to compute a pairwise Martin distance between each pair of trajectories, performing the code on distributed workers does not seem to be a good choice. Moreover, by considering the fact that we only need to compute these heavy $m \times{m}$ matrix computation once (when all the trajectories of different videos are extracted) we decided to set that section of our framework to work parallelly using python\textquotesingle   joblib parallelization libraries.

\subsection{Configuring the Dask pipeline}

We used a publicly available tool (dask-gke) provided by the Dask team to create and manage Dask clusters on GCP.
Dask-gke creates kubernetes-managed container clusters off blueprints detailing the number and type of kubernetes pods in the cluster: Dask scheduler, Dask worker, and/or a Jupyter notebook. Each type of pod can be spawned with a particular Docker image, depending on its purpose. We created one Docker image, equipped with all the frameworks and libraries required by our image processing pipeline (includes Python, scikit-learn, OpenCV, Dask, Jupyter, etc.) and used it for all the pods in the Dask cluster blueprint (yaml file).
Dask-gke can spawn clusters and reconfigure them as need (expand or shrink the number of pods and/or the instances backing them). Once the cluster is up, with all its pods, we can submit Dask jobs to the scheduler directly, or through the Jupyter service connecting to the scheduler\textquotesingle s internal address.
The Dask scheduler will manage the assignment and execution of the tasks with the workers as our python pipeline code executes.

\begin{figure}
\centerline{\includegraphics[width=9cm, height=5.5cm]{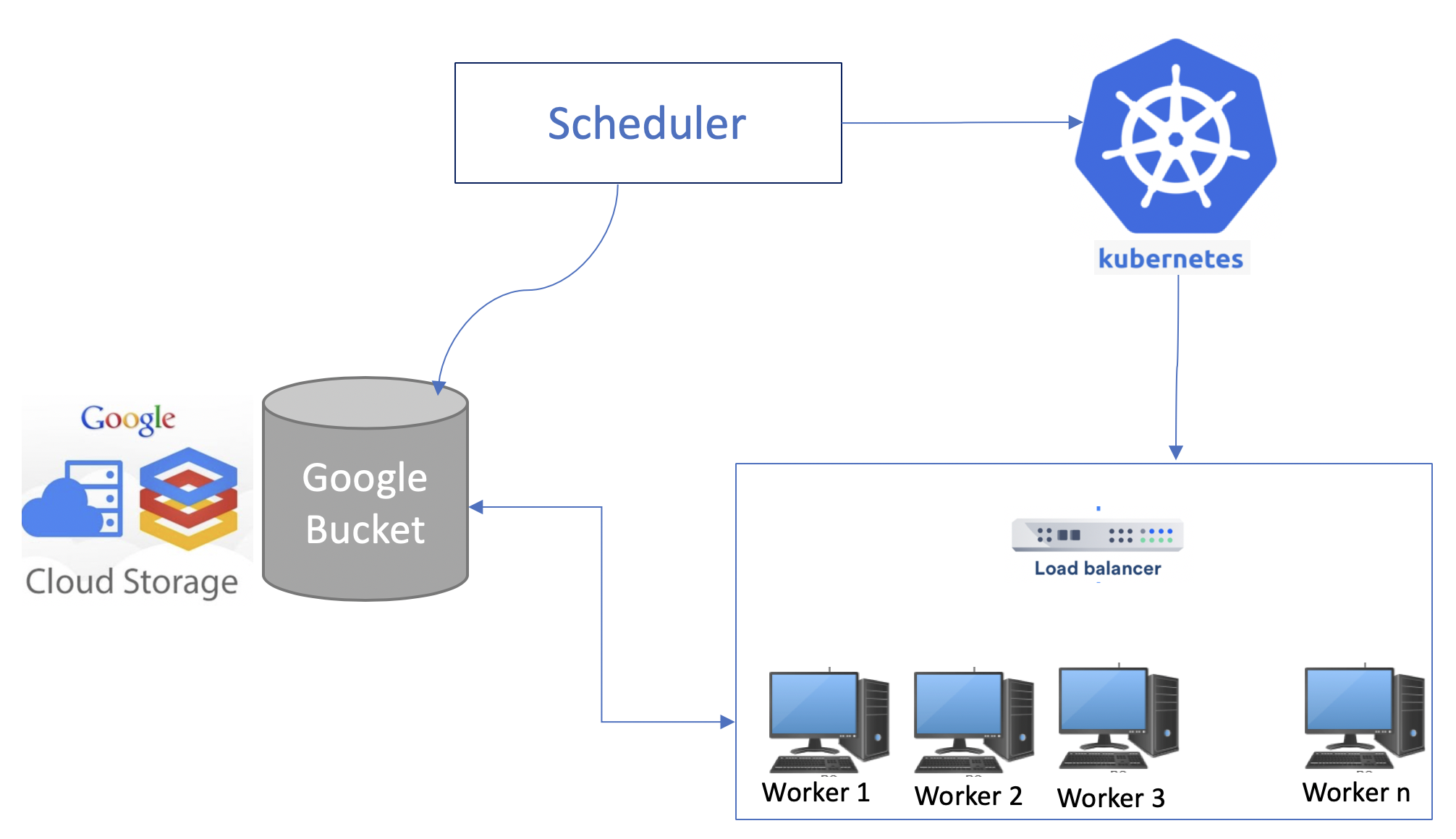}}
\caption{\textit{The schema of the GCP cluster on which we performed our distributed Dask version of our pipeline.}}
\label{fig:11}
\end{figure}
\subsection{Performance Comparison}
We ran the serial version of our code on a local machine with Intel \textit{Core i9} CPU and \textit{128}GB of RAM.  For the distributed Dask version of our code, we created \textit{2} clusters on GCP: one with \textit{3} workers, \textit{8} pods, and the second one with \textit{8} workers and \textit{21} pods.

For the parallelization part in the \textit{clustering section} \ref{clusteringsection} of our framework, we used Python\textquotesingle s joblib backend and the multiprocessing library. For this step, we used the same local \textit{Core i9} machine.

The computation needs here are not only dictated by the size of the dataset but also the number of cells inherent in the videos and the results of the sparsification. In other words, the level of sparsity plays a crucial role in our computational time: a greater number of cells results in more cell particles to extract, more 3D shapes to compute their centers and finally, more trajectories to compare. Thus, wall-time is the indicative measure of how efficient a pipeline is, for the same dataset. 

\begin{figure}
\centerline{\includegraphics[width=9cm, height=5.5cm]{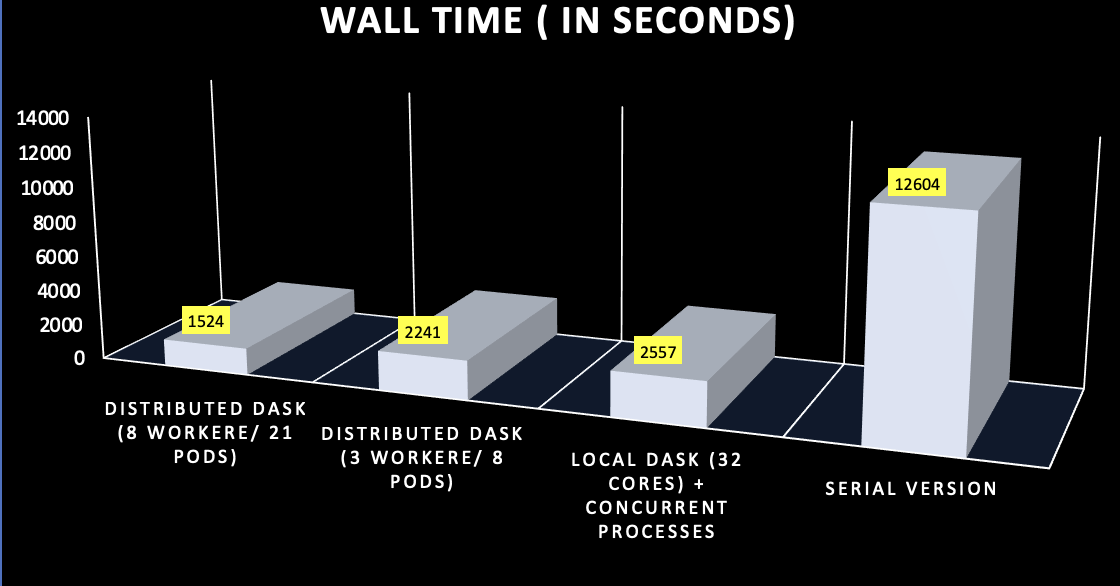}}
\caption{\textit{The comutational time of our platform in different settings: (right-to-left) Serial, Dask-python on a local machine, Dask-python on a small GCP cluster and finally Dask-python on a larger GCP cluster.}}
\label{fig:12}
\end{figure}

Fig. \ref{fig:12} shows that the wall time in a cluster of $8$ workers and $21$ pods has sharply decreased from $12,604$ seconds (serial version) to $1,524$ seconds.
In other words, the larger cluster achieved a 87.9\% reduction in wall-time compared to the serial version. The smaller cluster achieved 82.2\% reduction in wall-time. Lastly, the local Dask version achieved 79.7\% wall-time reduction compared to the serial version.
This shows the vast improvements in our pipeline achieved by incorporating task distribution and parallelization over serial processing.
Comparing the performance of the two clusters shows $32.0\%$ reduction in wall-time by using a larger cluster compared to the smaller one which proves the scalability of our pipeline.

\section{Conclusions and Future work}
In this paper, we proposed a scalable lightweight framework to track and analyze the motion structure of \textit{Toxoplasma gondii} in multiple 3D videos.  From a biological point of view, researchers suggested that there is only one type of motion in \textit{T. gondii} in 3D \cite{b14}. However, our research indicates that there are at least three types of motion patterns that we extracted and analyzed using our framework on 4D data. From a data science point of view, the research indicates that by applying a hybrid model consisting of computer vision techniques, large scale analysis and cutting-edge machine learning tools applied on massive data, we are able to assist biologists to discover new areas and hypotheses that are not easily possible to explore otherwise. Although our experiments were conducted with \textit{T. gondii} as its principal applicant; In theory, it should be applicable to other similar species of motile parasites, including \textit{Plasmodium ookinetes} (causative agent of malaria) and sporozoites that undergo similar motility patterns, as well as other numerous applications of motility tracking throughout public health that would benefit from this framework.

As future work, first, we aim to apply our framework on more datasets, primarily, we are going to apply it on experiments with the presence of pharmacological drugs that stimulate or inhibit the motion of the cells and analyze how these stimuli can change the motion of \textit{T. gondii}. Moreover, we aim to develop fully unsupervised deep learning-based methods, especially in place of the current distance metric and trajectory clustering. A deep network would allow us to create an end-to-end unsupervised preprocessing, tracking, parameterization, and clustering framework, with a data-driven distance learning metric. Right now, we rely on hand-tuned hyperparameters and distance metrics to perform our analysis. While this achieves decent results, it makes for a brittle model in terms of its configuration space. Including an attention mechanism in an end-to-end framework would allow the model to learn, in a fully unsupervised way, what parts pay attention to in the first place, rather than relying on us to properly tune the parameters.

\section*{Acknowledgments}
The authors acknowledge Dr. Kyle Johnsen and the Georgia Informatics Institutes for providing the computational resources for this project. This work is supported in part by the GII Fellowship. We gratefully acknowledge the support of NVIDIA Corporation with the donation of a Titan X Pascal GPU used in this research. We acknowledge and thank Google for a generous research grant to use on their Compute Platform. We acknowledge partial support from the NSF Advances in Biological Informatics (ABI) under award number $1458766$ and US Public Health Service grants $AI139201$ and $AI137767$.

\end{document}